\documentclass{article}

\usepackage{microtype}
\usepackage{graphicx}
\usepackage{overpic}
\usepackage{subfigure}
\usepackage{booktabs} % for professional tables

\usepackage{stfloats}

\usepackage{float}

\usepackage{hyperref}

\usepackage[accepted]{icml2024}

\usepackage{amsmath}
\usepackage{amssymb}
\usepackage{mathtools}
\usepackage{amsthm}

\usepackage[capitalize,noabbrev]{cleveref}

\usepackage{bm}
\usepackage{multirow}
\usepackage{lipsum}
\usepackage{placeins}
\usepackage{graphicx}
\usepackage{bm}
\usepackage{rotating}
\usepackage{multirow, makecell}

\DeclareMathOperator*{\dd}{d}
\newtheorem{thm}{Theorem}
\newcommand{\code}[1]{\texttt{#1}}

\newcommand\methodname{FOMA}

\theoremstyle{plain}

\theoremstyle{definition}

\theoremstyle{remark}

\usepackage[textsize=tiny]{todonotes}

\icmltitlerunning{First-Order Manifold Data Augmentation for Regression Learning}

\begin{document}

\twocolumn[
\icmltitle{First-Order Manifold Data Augmentation for Regression Learning}



\icmlsetsymbol{equal}{*}

\begin{icmlauthorlist}
\icmlauthor{Ilya Kaufman}{sch}
\icmlauthor{Omri Azencot}{sch}
\end{icmlauthorlist}

\icmlaffiliation{sch}{Department of Computer Science, Ben-Gurion University of the Negev, Beer-Sheva, Israel}
\icmlcorrespondingauthor{Ilya Kaufman}{ilyakau@post.bgu.ac.il}
% \icmlcorrespondingauthor{Omri Azencot}{azencot@cs.bgu.ac.il}

\icmlkeywords{Machine Learning, ICML}

\vskip 0.3in
]



\printAffiliationsAndNotice{}  % leave blank if no need to mention equal contribution

\begin{abstract}        
    Data augmentation (DA) methods tailored to specific domains generate synthetic samples by applying transformations that are appropriate for the characteristics of the underlying data domain, such as rotations on images and time warping on time series data. In contrast, \emph{domain-independent} approaches, e.g. \code{mixup}, are applicable to various data modalities, and as such they are general and versatile. While regularizing classification tasks via DA is a well-explored research topic, the effect of DA on regression problems received less attention. To bridge this gap, we study the problem of domain-independent augmentation for regression, and we introduce \code{\methodname}: a new data-driven domain-independent data augmentation method. Essentially, our approach samples new examples from the tangent planes of the train distribution. Augmenting data in this way aligns with the network tendency towards capturing the dominant features of its input signals. We evaluate \code{\methodname} on in-distribution generalization and out-of-distribution robustness benchmarks, and we show that it improves the generalization of several neural architectures. We also find that strong baselines based on \code{mixup} are less effective in comparison to our approach. Our code is publicly available at \url{https://github.com/azencot-group/FOMA}.
\end{abstract}

\section{Introduction}

% classification vs regression; regularization and overparameterization
Classification and regression problems primarily differ in their output's domain. In classification, we have a finite set of labels, whereas in regression, the range is an infinite set of quantities---either discrete or continuous. In classical work~\citep{devroye2013probabilistic}, classification is argued to be ``easier'' than regression, but more generally, it is agreed by many that classification and regression problems should be treated differently~\citep{muthukumar2021classification}. Particularly, the differences between classification and regression are actively explored in the context of regularization. Regularizing neural networks to improve their performance on new samples has received a lot of attention in the past few years. One of the main reasons for this increased interest is that most of the recent successful neural models are \emph{overparameterized}. Namely, the amount of learnable parameters is significantly larger than the number of available training samples \citep{allen2019learning, allen2019convergence}, and thus regularization is often necessary to alleviate overfitting issues. Recent studies on overparameterized linear models identify conditions under which overfitting is ``benign'' in regression \citep{bartlett2020benign}, and uncover the relationship between the choice of loss functions in classification and regression tasks~\citep{muthukumar2021classification}. Still, the regularization of deep neural regression networks is not well understood.

% data augmentation; DA and linear regression; gap
In this work, we focus on regularizing deep models via Data Augmentation (DA), where data samples are artificially generated and used during training. In general, DA techniques can be categorized into domain-dependent (DD) methods and domain-independent (DI) approaches. The former are specific for a certain data modality such as images or text, whereas the latter typically do not depend on the data format. Numerous DD- and DI-DA approaches are available for classification tasks~\citep{shorten2019survey, shorten2021text}, and many of them consistently improve over non-augmented models. Unfortunately, DI-DA for regression problems is underexplored. Recent works on linear models study the connection between the DA policy and optimization~\citep{hanin2021data}, as well as the generalization effects of linear DA transformations~\citep{wu2020generalization}. We contribute to this line of work by proposing and analyzing a new domain-independent data augmentation method for nonlinear deep regression, and by testing our approach on in-distribution generalization and out-of-distribution robustness tasks~\cite{yao2022cmix}.

% DA for classification, ERM vs. VRM, does not work for regression
Many strong data augmentation methods were proposed in the past few years. Particularly relevant to our study is the family of \code{mixup}-based techniques that are commonly used in classification applications. The original method, \code{mixup}~\citep{zhang2017mixup}, produces convex combinations of training samples, promoting linear behavior for in-between samples. The method is domain-independent and data-agnostic, i.e., it is indifferent to the given data samples. \code{Mixup} was shown to solve the Vicinal Risk Minimization (VRM) problem instead of the usual Empirical Risk Minimization (ERM) problem. In comparison, our approach can also be viewed as solving a VRM problem, and it is domain-independent and \emph{data-driven}, namely, augmentations depend on the given data distribution. In our extensive evaluations, we will show that \code{mixup}-based methods are less effective for regression in comparison to our approach.

% types of NN regression, scarcity of domain-independent DA approaches for regression

% contributions: domain-independent, data-driven, incorporated with other approaches, relation to noise injection, experimental setup, 
\paragraph{Contributions.} Challenged by the differences between classification and regression and motivated by the success of domain-independent methods such as \code{mixup}, we propose a simple, domain-independent and data-driven DA routine, termed First-Order Manifold Augmentation (\code{\methodname}). Let $X, Y$ be the input and output mini-batch tensors, respectively, and let $Z_{l} = g_l(X)$ be the hidden code at layer $l$. Our method produces new training samples $Z_{l}(\lambda), Y(\lambda)$ from the given ones by scaling down their small singular values by a random $\lambda \in [0, 1]$. At its core, \code{\methodname} incorporates into training the assumption that data with similar dominant components of the train set should be treated as true samples. Our implementation of \code{\methodname} is fully differentiable, and thus it is applicable to any layer of a given network.

% paper organization: timings, ablation, why mixup fails/decision boundary
% TODO: update based on results and appendix
We detail \code{\methodname} in Sec.~\ref{sec:method}, motivating our design choices and illustrating its effect on data. We analyze our approach using perturbation theory and introduce its associated vicinal risk minimization (Sec.~\ref{sec:analysis}). Our experimental evaluation focuses on in-distribution generalization (Sec.~\ref{sec:id_gen}) and on out-of-distribution robustness (Sec.~\ref{sec:ood_gen}), where we empirically demonstrate the superiority of \code{\methodname}. We offer a potential explanation to the success of our method (Sec.~\ref{sec:method}, App.~\ref{app:models_lamv}). Finally, an ablation study is performed, justifying our design choices (Sec.~\ref{subsec:abl}).

% TODO: motivate regression DA (why DA, why regression, why existing insufficient), motivate FOMA, 

\section{Related Work}

Deep neural networks regularization is an established research topic with several existing works \citep{goodfellow2016deep}. Common regularization approaches include weight decay, dropout \citep{srivastava2014dropout}, batch normalization \citep{ioffe2015batch}, and data augmentation (DA). Here, we categorize DA techniques to be either domain-dependent or domain-independent. Domain-dependent DA was shown to be effective for, e.g., image data \citep{lecun1998gradient} and audio signals \citep{park2019specaugment}, among other domains. However, adapting these methods to new data formats is typically challenging and often infeasible. While several works focused on automatic augmentation~\cite{lemley2017smart, cubuk2019autoaugment, lim2019fast, tian2020improving, cubuk2020randaugment}, there is concurrently an increased interest on domain-independent DA methods, allowing to regularize neural networks when only basic data assumptions are allowed~\cite{naiman2023sample}. We focus in what follows on \emph{domain-independent} techniques that were proposed in the context of classification and regression problems.

\paragraph{DA for classification.} Zhang et al. (\citeyear{zhang2017mixup}) proposed to perform convex mixing of input samples as well as one-hot output labels during training. The new training procedure, named \code{mixup}, minimizes the Vicinal Risk Minimization (VRM) problem instead of the typical Empirical Risk Minimization (ERM). Many extensions of \code{mixup} were proposed, including mixing latent features~\citep{verma2019manifold}, same-class mixing~\citep{devries2017dataset}, among other extensions~\citep{guo2019mixup, hendrycks2019augmix, yun2019cutmix, berthelot2019mixmatch, greenewald2021k, lim2021noisy}. ISDA~\cite{wang2019implicit} formulates a new cross-entropy loss for DA-based training using the per-class covariance matrix.

\paragraph{DA for regression.} Significantly less attention has been drawn to designing domain-independent data augmentation for regression tasks. A recent survey~\citep{wen2020time} on DA for time series data lists a few basic augmentation tools. Dubost et al. (\citeyear{dubost2019hydranet}) propose to recombine samples for regression tasks with countable outputs, and thus their method can not be directly extended to the uncountable regime. \code{RegMix} \citep{hwang2021regmix} developed a meta learning framework based on reinforcement learning for mixing samples in their neighborhood. A recent work~\citep{yao2022cmix} showed that applying vanilla mixup with adjusted sampling probabilities based on label similarity can improve generalization on regression tasks. Another work~\citep{schneider2023anchor} suggests DA based on Anchor regression~\cite{rothenhausler2021anchor} which allows mixing multiple samples based on their cluster that encodes a homogeneous group of observations.

\section{First-Order Manifold Augmentation}
\label{sec:method}

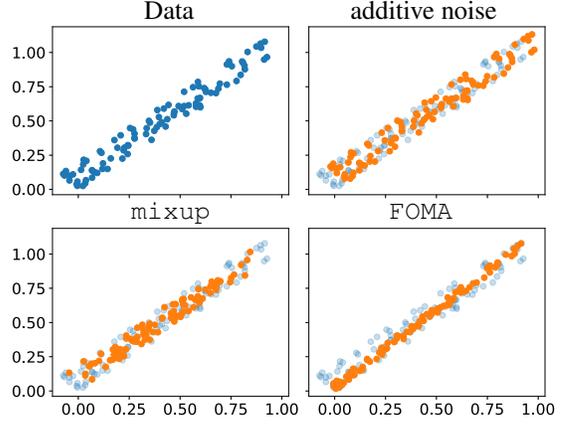
\begin{figure*}[t]
  \centering
    \begin{overpic}[width=1\linewidth]{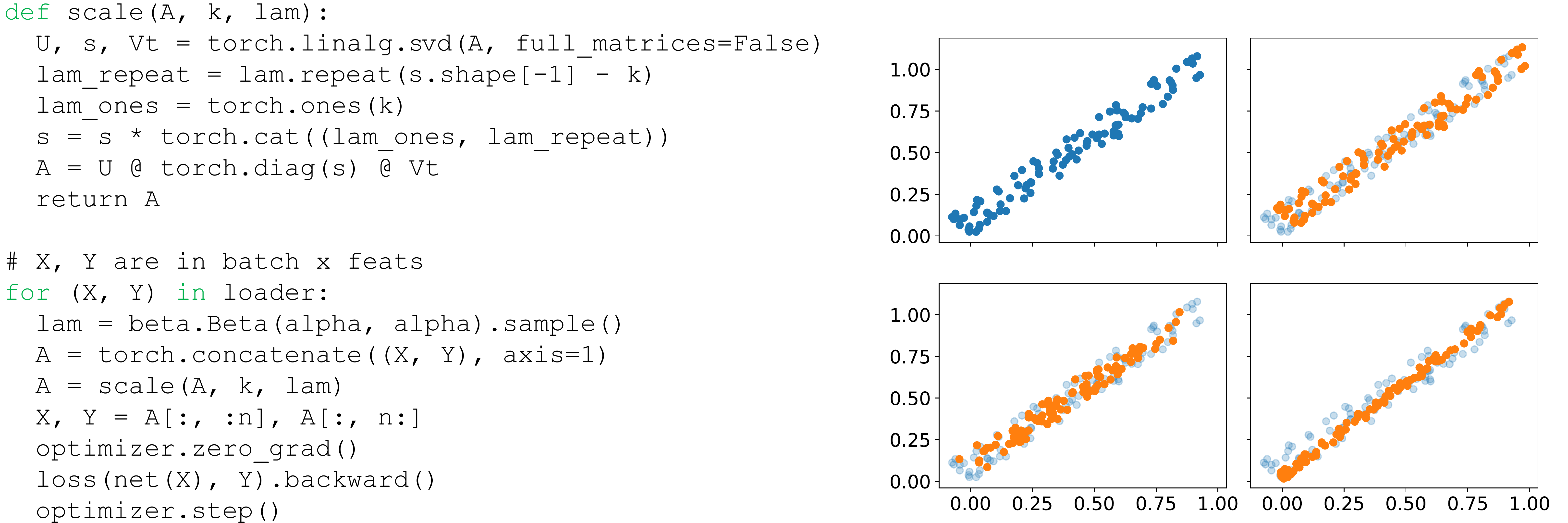} 
        \put(67,31.5){Data}
        \put(83,31.5){additive noise}
        \put(66,16){\code{mixup}}
        \put(86,16){\code{\methodname}}
    \end{overpic}
  \caption{We show the pseudocode for \code{\methodname} at the input level, $l=0$ (left). We demonstrate the effect of a few DA methods on 2D data whose intrinsic dimension is one (right).}  
  \label{fig:code_illus}
\end{figure*}

% problem formulation: regression, extend training distribution, 
A learning task is typically described as a function which maps inputs to outputs. In this view, a learning model is approximating that function using e.g., a neural network, and it is formulated via $f: \mathcal{X} \rightarrow \mathcal{Y}$, denoting the input and output domains by $\mathcal{X}$ and $\mathcal{Y}$, respectively. A regression problem is such that the output domain is (un)countable, e.g., $\mathcal{Y} \subset \mathbb{N}^m$ or $\mathcal{Y} \subset \mathbb{R}^m$. We consider the general setting where $\mathcal{X}\subset \mathbb{R}^n, \mathcal{Y} \subset \mathbb{R}^m$. During training, the learning model is provided with a training set $\mathcal{D} = \{ (x_i, y_i)\}_{i=1}^N $, sampled from $(x_i, y_i) \sim \mathcal{P}$. Our method extends $\mathcal{D}$ by producing a new training distribution as we describe below.

% our approach: SVD, scale small singular values
To generate new samples, we consider the singular value decomposition (\code{SVD}) of a matrix $M$ $\in \mathbb{R}^{q \times r}, q \geq r$ which is given by  $M$ $= U S V^T$. $U, V$ are orthogonal, and $S$ is diagonal consisting of the singular values ordered by $\sigma_1 \geq \sigma_2 \geq \dots \geq \sigma_r \geq 0$. \code{SVD} is intimately related to principal component analysis (\code{PCA}) which in turn is heavily studied in manifold learning and dimensionality reduction~\citep{ma2012manifold}. It is well known that the best rank $k$ approximation of $M$ is given by omitting the last $(r-k)$ singular values, i.e., $M_k = \sum_{j=1}^k \sigma_j u_j v_j^T$~\citep{eckart1936approximation}. The matrix $M_k$ preserves the ($k$) dominant components in $M$, and discards the rest. Further, \code{SVD} is also known to yield a first-order approximation of the data manifold~\cite{singer2012vector, kaufman2023data}. Our key insight is that scaling the small singular values produces training samples that are in close proximity to the manifold, and thus to the true data distribution $\mathcal{P}$.

In particular, based on the manifold hypothesis~\cite{cayton2008algorithms}, we assume that the data samples $\mathcal{D}$ live on or close to a manifold $\mathcal{M} \subset \mathcal{X} \times \mathcal{Y}$. We denote by $\mathcal{T}(x,y)$ the tangent plane of the data manifold $\mathcal{M}$ at the point $(x, y) \in \mathcal{M}$. Namely, $\mathcal{T}(x, y)$ is the linear approximation of $\mathcal{M}$ at $(x,y)$. Below, we utilize \code{SVD} to approximate $\mathcal{T}(x, y)$, and to generate artificial samples by considering pairs $(\tilde{x}, \tilde{y})$ in the tangent plane of the $(x, y)$.

% pseudo code and illustration of method
% Let the input and output mini-batch tensors $X \in \mathbb{R}^{b \times n}$ and $Y \in \mathbb{R}^{b \times m}$, respectively, where w.l.o.g $b \geq n+m$ is the batch size. We denote the network by $f(X) = f_l (g_l(X))$, $Z_l := g_l(X)$ where $g_l$ maps inputs to latent representations $Z_l$ at layer $l \in [0, L]$, and $f_l$ maps latent vectors to outputs. Let $\lambda \sim \text{Beta}(\alpha, \alpha)$ for $\alpha \in (0, \infty)$, then, the new samples $Z_l(\lambda), Y(\lambda)$ are defined via
% \begin{align*}
%     & A := [Z_l, \; Y] = U S V^T \in \mathbb{R}^{b \times \left ( n+m \right)} \ , \\
%     & A(\lambda) := [Z_l(\lambda), \; Y(\lambda)] = U S(\lambda) V^T \ ,
% \end{align*}
% where $[\cdot, \cdot]$ concatenates along columns, and $S(\lambda)$ is the diagonal matrix of scaled down singular values. Namely, we compute $S(\lambda) = \text{diag}(\sigma_1, \dots, \sigma_d \; | \; \lambda \sigma_{d+1}, \dots, \lambda \sigma_{n+m})$. The value $d$ is the intrinsic dimension (ID) of $A$, i.e., the minimal number of features needed to represent the data with little information loss~\cite{facco2017estimating}. The ID can be estimated for the entire dataset at the input level, or alternatively, approximated per batch at any layer $l\in[0,L]$.

Let the input and output mini-batch tensors $X \in \mathbb{R}^{b \times n_{0}}$ and $Y \in \mathbb{R}^{b \times m}$, respectively, where w.l.o.g $b \geq n_{0}+m$ is the batch size. We denote the network by $f(X) = f_l (g_l(X))$, $Z_l := g_l(X)$ where $g_l$ maps inputs to latent representations $Z_l$ $\in \mathbb{R}^{b \times n_{l}}$ at layer $l \in [0, L]$, and $f_l$ maps latent vectors to outputs. Let $\lambda \sim \text{Beta}(\alpha, \alpha)$ for $\alpha \in (0, \infty)$ and $k \in [1, n_{l}+m]$ be the index of the singular value after which we scale down. Then, the new artificial samples $Z_l(\lambda, k), Y(\lambda, k)$ are defined via
\begin{align*}
    & A := [Z_l, \; Y] = U S V^T \in \mathbb{R}^{b \times (n_{l}+m)} \ , \\
    & A(\lambda, \, k)  := U S(\lambda, \, k) V^T \in \mathbb{R}^{b \times (n_{l}+m)}  \ , \\
    & Z_l(\lambda, \, k) := A(\lambda, \, k)_{1:n_{l}}   \in \mathbb{R}^{b \times n_{l}} \ , \\
    & Y(\lambda, \, k) := A(\lambda, \, k)_{n_{l}+1:n_{l}+m} \in \mathbb{R}^{b \times m}  \ ,
\end{align*}

where $[\cdot, \cdot]$ concatenates along columns, $A_{1:i}$ stands for the first $i$ column vectors in $A$, and $S(\lambda, \, k)$ is the diagonal matrix of scaled down singular values. Namely, we compute $S(\lambda, \, k) = \text{diag}(\sigma_1, \dots, \sigma_k \; | \; \lambda \sigma_{k+1}, \dots, \lambda \sigma_{n_{l}+m})$. We propose two methods for choosing the parameter $k$, one is based on the intrinsic dimension of the data, denoted by \code{\methodname} and the second is based on the explained variance of the data denoted by \code{\methodname}$_{\rho}$.

\begin{table*}[!b]
\caption{Results for different batch selection methods \code{\methodname}.}
\label{tab:b_selection}
\centering
\vskip 0.1in

    \begin{tabular}{l|c|c|c|c|c||c|c|c|c}
    \toprule 
    Dataset & Airfoil\,$\downarrow$ & NO2\,$\downarrow$ & Exchange\,$\downarrow$ & Electricity\,$\downarrow$ & Echo\,$\downarrow$ & RCF\,$\downarrow$ & Crimes\,$\downarrow$ & Poverty\,$\uparrow$ & DTI\,$\uparrow$ \\
    \midrule 
    \methodname - random& 2.159 & \textbf{0.515} & 0.014 & 0.059 & 5.248 &0.175 & 0.144 & 0.433 & 0.491 \\
    \methodname - close& \textbf{1.646} & 0.521 & \textbf{0.013} & \textbf{0.058} & \textbf{5.215} &\textbf{0.171} & \textbf{0.132} & \textbf{0.492} & \textbf{0.496} \\
    \midrule 
    \methodname $\rho$ - random & 2.012 & 0.514 & 0.014 & 0.060 & \textbf{5.224} & 0.163 &0.134 & 0.409 & \textbf{0.508}  \\
        \methodname $\rho$ - close & \textbf{1.471} & \textbf{0.512} & \textbf{0.013} & \textbf{0.059} & 5.512 &\textbf{0.159} & \textbf{0.128} & \textbf{0.488} &0.503 \\
    \bottomrule
    \end{tabular}
    \vskip -0.1in

\end{table*}

\textbf{Intrinsic dimension.} We set the value of $k$ to be equal the intrinsic dimension (ID) of $A$, i.e., the minimal number of features needed to represent the data with little information loss~\cite{facco2017estimating}. The ID can be estimated for the entire dataset at the input level, or alternatively, approximated per batch during the training process. Setting $k$ to equal the ID is motivated by our analysis below in Sec.~\ref{sec:analysis}, where we show that \code{\methodname} is equivalent to sampling from the tangent space of the manifold. In practice, we use $k$ dominant singular vectors to approximate the tangent space. 

\begin{figure*}[t]
  \centering
  \begin{overpic}[width=1\linewidth]{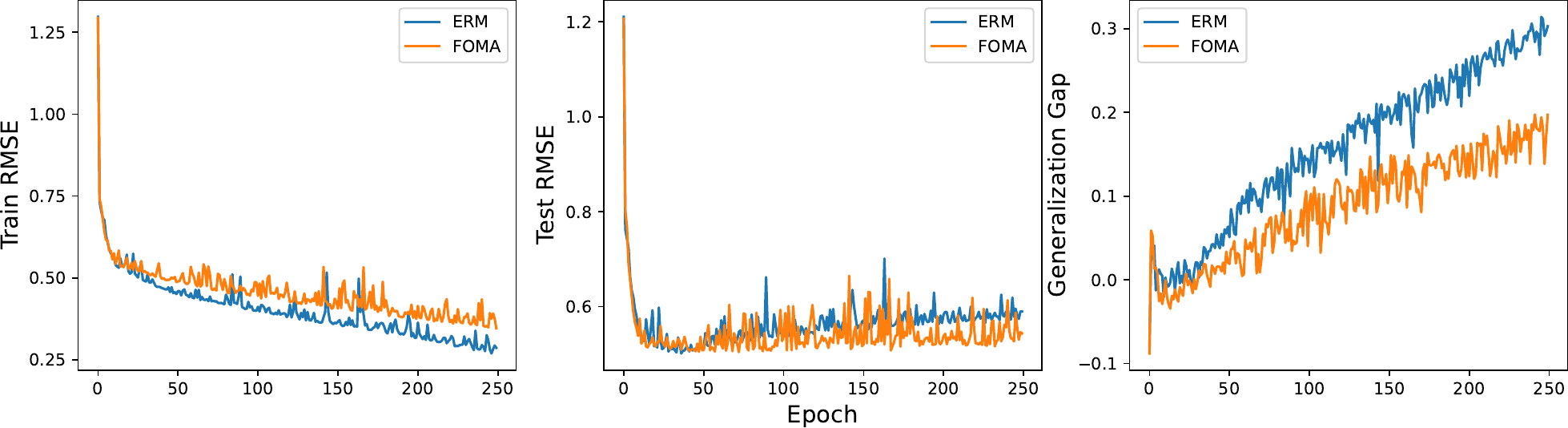}
      \put(17, 25){(a)} \put(51,25){(b)} \put(85, 25){(c)}
  \end{overpic}
  \caption{Training stability and overfitting. (a) RMSE loss on the train set. (b) RMSE loss on the test set. (c) Generalization gap: the difference between test error and train error}
  \label{fig:stability_app}
\end{figure*}

\textbf{Explained variance.} In addition to the ID, we also consider the value $k$ to depend on the hyper-parameter $\rho \in [0, 1]$ that represents the ``amount'' of signal to keep unchanged, i.e., 
\begin{align*}
    k = \arg\max_{\tilde{k}} \sum_{j=1}^{\tilde{k}} \sigma_j / \sum_j \sigma_j \leq \rho \ .
\end{align*}
Similar to \code{mixup}~\citep{zhang2017mixup}, our method recovers the original dataset $\mathcal{D}$ as $\alpha \rightarrow 0$, $\forall k$.

% loss function, implementation, and design choices
The loss function associated with \code{\methodname} is
\begin{align*}
    \mathcal{L}(f) &= \mathbb{E}_{(X,Y)} \, \mathbb{E}_{\lambda} \, \mathbb{E}_{l} \, c_\lambda \left[ (f_l, 1) \circ \chi(g_l(X), Y)) \right] \ , \\
    \text{s.t.} & \quad (X, Y) \sim \mathcal{D},  \; \lambda \sim \mathcal{H}(\sigma), \; l \sim [0, L] \ ,
\end{align*}
where $c_\lambda: \mathbb{R}^{b \times m} \times \mathbb{R}^{b \times m} \rightarrow [0, \infty)$ is a cost function, typically mean squared error (MSE). The \code{scale} transform $\chi$ takes a pair of tensors $g_l(X), Y$, and it scales down by $\lambda$ the last $(n_l + m - k)$ singular values of their concatenation. A key attribute of \code{\methodname} is that it is fully \emph{differentiable} since the singular value decomposition can be backpropagated~\citep{ionescu2015matrix}. We provide an example PyTorch pseudocode in Fig.~\ref{fig:code_illus} (left). The computational complexity of \code{\methodname} is governed by \code{SVD} calculation which has a complexity of $\mathcal{O}(\min(qr^2,rq^2))$ for a $q\times r$ matrix.

\begin{figure*}[t]
  \centering
  \includegraphics[width=1\linewidth]{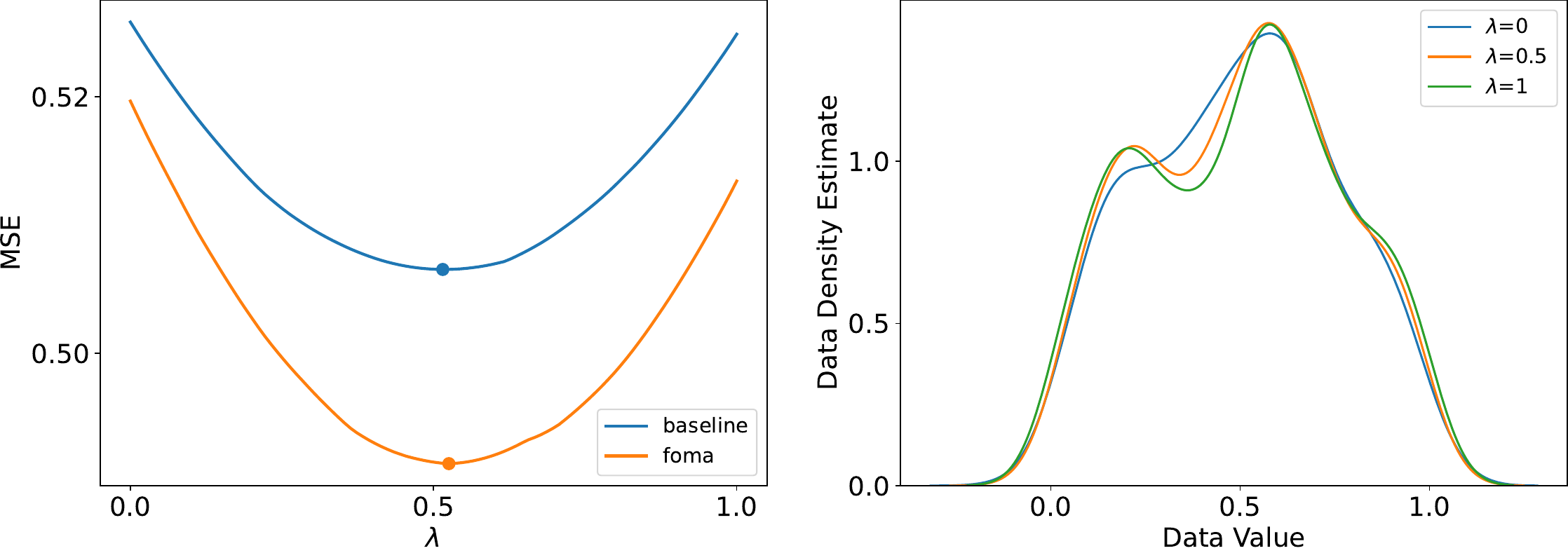}
  \caption{Evaluating a non-augmented model and a model trained with \code{\methodname} on train data whose small singular values are scaled down for different values of $\lambda$ (left). We show on the right panel the probability density function of the original data (green), and its modifications using $\lambda=0$ (blue), and $\lambda=0.5$ (orange).}  
  \label{fig:cz_analysis}
\end{figure*}

\paragraph{Design choices.} For certain $\lambda$ values, the new sample $Z_l(\lambda, k), Y(\lambda, k)$ may be too far from $\mathcal{P}$. With this in mind, we explored the option of scaling down the loss function $c(\cdot, \cdot)$ by a parameter $\mu(\lambda)$ in addition to modifying the singular values. However, we tested various profiles $\mu(\lambda)$ and discovered the most consistent models are obtained when no scaling of loss occurs, see Sec.~\ref{subsec:abl}. Importantly, this means that our approach adopts a different ansatz in comparison to \code{mixup}-based methodologies. While \code{mixup} incorporates uncertainty into the model training using ``in-between'' samples and labels, our method uses the new data as if it was sampled from the true distribution, since we do not scale $c$. An alternative option which would be conceptually closer to \code{mixup} is to scale the \emph{large} singular values as well as the loss term. We show in Sec.~\ref{subsec:abl} that this choice is usually inferior to \code{\methodname}.

\paragraph{Batch selection.} Given a sample $(x,y)\in \mathcal{X} \times \mathcal{Y}$, we aim to produce training samples that are in close proximity to the manifold  $\mathcal{M}$, and thus to the true data distribution $\mathcal{P}$. However, we do not know the structure of $\mathcal{M}$, and thus we can not sample from it directly, unfortunately. To overcome this challenge, we sample from a linear approximation of the $\mathcal{M}$ at $(x,y)$ given by the tangent plane $\mathcal{T}(x,y)$. In practice, we need a set of points $N_{(x,y)}=\{(x_i,y_i )\}$ in the neighborhood of $(x,y)$ to estimate $\mathcal{T}(x,y)$ via \code{SVD}. The proximity of the points $N_{(x,y)}$ to $(x,y)$ has a direct effect on the quality of the approximation of the tangent plane. In practice, we construct batches of points that are close to each other as measured by the Euclidean distances between the labels in $\mathcal{Y}$. We hypothesize that better approximations should generate samples closer to the data manifold, improving empirical test results. To validate our claim, we tested two batch selection methods for training 1) randomly selecting batches (random); and 2) constructing batches using samples of points that are close to each other (close). For each setting we trained several models and selected the model that achieved the best performance on the validation set. In Table~\ref{tab:b_selection}, we report the results of batch selection methods on the test set. It is notable that constructing batches of close points outperforms random batch selection. 
% effect of selecting random batch vs constructing it from close points

\paragraph{Stability and overfitting}
We trained two models on the NO2 dataset, one without using our method (ERM) and one with our method (FOMA) and report the results in Fig.~\ref{fig:stability_app}. We observe that (a) the training process of \code{\methodname} is stable with respect to ERM, i.e., the loss is monotonically decreasing with relatively small fluctuations. Furthermore, \code{\methodname} exhibits better test performance (b) and smaller generalization gap (c).

\paragraph{Computational resources}
\label{app:speed}
The computational complexity of \code{\methodname} is governed by \code{SVD} calculation which has a complexity of $\mathcal{O}(\min(qr^2,rq^2))$ for a $q\times r$ matrix. In Table~\ref{tab:speed_foma}, we compare the average epoch time across 50 epochs in seconds to provide an estimate for the empirical computational cost of \code{\methodname}. The results are obtained with a single \code{RTX3090} GPU. Note that the runtime of \code{\methodname} is very dependant on the batch size. Larger batch size results in less \code{SVD} computations, and depending on implementation, \code{SVD} of larger matrices can be evaluated faster.
 % table for foma

\begin{table}[ht]
\caption{Training times using \code{\methodname} (in seconds).}
\label{tab:speed_foma}
\centering
\begin{small}
\begin{sc}
\vskip 0.1in
    \begin{tabular}{l|ccc}
    \toprule 
    Dataset & Airfoil & Electricity & DTI \\
    \midrule 
    ERM & 0.058 & 1.406& 33.374\\
    
    \methodname - input & 0.148 & 3.101& 43.433 \\
    \methodname - latent & 0.286 & 2.837& 51.399\\
    \methodname - both & 0.321 & 4.376& 63.071 \\

    \methodname $\rho$ - input & 0.119 & 2.806& 41.527 \\
    \methodname $\rho$ - latent & 0.149  & 3.021& 43.985\\
    \methodname $\rho$ - both & 0.262 & 4.315& 57.35 \\
    \bottomrule
    \end{tabular}
    \vskip -0.1in
\end{sc}
\end{small}
\end{table}

% what FOMA is doing?
\paragraph{The effect of \code{\methodname} on data and learning.} We generated a 2D point cloud whose intrinsic dimension is one (shown in blue, Fig.~\ref{fig:code_illus}), and we applied different DA methods on this data. The three panels in the figure show in orange the augmented data when using additive noise, \code{mixup}, and \code{\methodname} with $\lambda=0.5$ and  over the original point cloud colored in light blue. Injecting noise alters each point in its neighborhood, whereas \code{mixup} draws the points towards the center of their convex hull. In contrast, \code{\methodname} aligns the new samples along the dominant component of the original data. Notably, our approach may increase the span of training data, and thus it can improve estimation in regression as was recently shown in~\citep{wu2020generalization}.

% IMPORTANT: Fig. 2 left uses ALL the data, since the test set is extremely small
We argue that training on samples created with our method encourages the inherent tendency of the network to model the dominant parts of the data better~\cite{naiman2023operator}. To demonstrate this phenomenon, we trained a three-layer fully connected network with and without \code{\methodname} on the NO2 dataset. The trained models are evaluated on the test dataset modified using a $100$ varying $\lambda \in [0, 1]$ values, see Fig.~\ref{fig:cz_analysis} (left). Namely, we modify the singular values of every batch in the dataset for each $\lambda$, and feed the resulting data for inference. Surprisingly, the non-augmented model (blue curve) performs \emph{better} on the unseen modified samples, yielding the minimum at $\lambda \approx 0.5$. Further, we note that for the majority of $\lambda$ values, the test MSE is lower than the error obtained for the original data $\mathcal{D}$ (i.e., for $\lambda=1$). In comparison, the regularized network attains a qualitatively similar plot in terms of the minimizing $\lambda$ and test MSE profile, however, the MSE is lower for all $\lambda$. This example shows that the (non-augmented and augmented) models generalize better to data projected to the manifold $\mathcal{M}$, except for a few low $\lambda$ values. We conclude that deep regression models may benefit from altering their training procedure by using samples closer to the data manifold $\mathcal{M}$. Finally, this behavior was found to be consistent across several architectures and datasets, see App.~\ref{app:models_lamv}.

Inspecting the data distribution and its modifications, reveals the differences between the data generated when select $\lambda$ values are used. The distribution of the original data (green) shown in Fig.~\ref{fig:cz_analysis} (right) is bimodal. In comparison, the blue curve ($\lambda=0$) for which the small singular values change to zeros, has a unimodal distribution. The orange curve ($\lambda=0.5$) for which both the non-augmented model and a model trained with \code{\methodname} attained the minimum loss, has a smoother transition between the major mode and the minor mode. From the analysis above, we conclude the following. First, the network prefers data whose distribution is simpler (orange) than the original distribution (blue), yet not too simple (green). Second, our regularization encourages this tendency by providing the model with such data, leading to improved MSE profiles. To the best of our knowledge, the above analysis is novel on deep regression models. 

Notably, while it may argued that the behavior in Fig.~\ref{fig:cz_analysis} (left) is natural and intuitive as the model ``simply'' performs better on denoised signals, we argue differently. In particular, this plot somewhat contradicts our understanding of overfitting which occurs in high probability for tiny datasets such as NO2 (the entire dataset is composed of $500$ entries) using multiple weights network such as the fully connected network we used with around $20k$ parameters. Specifically, since the data is highly likely to be overfit by the network, we expect the MSE value to be lowest for $\lambda=1$, and MSE value equal or higher for any $\lambda < 1$. Thus, we advocate that the above analysis may reveal a characteristic feature of regression neural networks. Our analysis is reinforced further as other datasets and architectures follow a similar pattern (App.~\ref{app:models_lamv}). Importantly, we are unaware of a similar experiment in the literature of deep regression neural networks.

\begin{table*}[!t]
\centering
\caption{Comparison of in-distribution generalization tasks. Bold values represent the best results and underlined values are second best. We report the average RMSE and MAPE over three seeds. Full results with standard deviation can be found in App~\ref{app:res_std}.}
\label{tab:id}
\vskip 0.1in
\resizebox{\textwidth}{!}{
\begin{tabular}{l|c|c|c|c|c|c|c|c|c|c}
\hline & \multicolumn{2}{|c}{ Airfoil } & \multicolumn{2}{|c}{ NO2 } & \multicolumn{2}{|c}{ Exchange-Rate } & \multicolumn{2}{|c}{ Electricity } & \multicolumn{2}{|c|}{ Echo }\\
\cline { 2 - 11 }& RMSE $\downarrow$ & MAPE (\%)$\downarrow$ & RMSE $\downarrow$ & MAPE (\%)$\downarrow$ & RMSE $\downarrow$ & MAPE (\%)$\downarrow$ & RMSE $\downarrow$ & MAPE (\%)$\downarrow$ & RMSE $\downarrow$ & MAPE (\%)$\downarrow$\\
\hline 
ERM$^{\dagger}$ & 2.901 & 1.753 & 0.537 &  13.615 & 0.0236 & 2.423 & 0.0581  & 13.861 & 5.402 & \underline{8.700}\\
Mixup$^{\dagger}$ & 3.730 & 2.327 & 0.528 & 13.534 & 0.0239 & 2.441 & 0.0585 & 14.306 & 5.393 & 8.838\\
Mani Mixup$^{\dagger}$ & 3.063 & 1.842 & 0.522 & 13.382 & 0.0242 & 2.475 & 0.0583 & 14.556 & 5.482 & 8.955\\
C-Mixup$^{\star}$ & 2.748 & 1.645 & 0.516 & \underline{13.069} & 0.024 & 2.456 & $\mathbf{0.057}$ & $\mathbf{13.349}$ &\underline{5.362} & 8.868 \\
ADA$^{\star}$ & 2.357 & 1.377 & $\underline{0.515}$ & 13.128 & 0.022 & 2.250 & 0.059 & 13.58 & - & -\\
\hline
\methodname & $\underline{1.646 }$ & $\underline{0.963 }$ & $0.521 $ & $13.23$ & $\underline{0.013}$ & $\underline{1.280}$ & $\underline{0.058}$ & 14.614 &$\mathbf{ 5.215}$& $\mathbf{8.331}$ \\
\methodname$_{\rho}$ & $\mathbf{1.471}$ & $\mathbf{0.816}$ & $\mathbf{0.512 }$ & $\mathbf{12.894}$ & $\mathbf{0.013 }$ & $\mathbf{1.262}$ & $0.059$ & \underline{13.437} & 5.512 & 8.742 \\
\hline
\hline 
C-Mixup$^{\dagger}$ & 2.717 & 1.610 & 0.509 & 12.998 & 0.020 & 2.041 & 0.057 & 13.372 &5.177 &8.435\\
ADA$^{\dagger}$ & 2.360 & 1.373 & 0.515 & 13.128 & 0.021 & 2.116 & 0.059 & 13.464 & - & -\\
\hline
\end{tabular}
}
\end{table*}

\section{Analysis}
\label{sec:analysis}

% the relation to perturbation
\paragraph{Relation to additive noise.} In what follows, we would like to answer the following question: Does applying \code{\methodname} is merely a variant of injecting additive noise? To this end, we analyze \code{\methodname} from a perturbation theory viewpoint. Specifically, we would like to understand how a random data perturbation affects the singular values of the data matrix $A \in \mathbb{R}^{q \times r}, \, q \geq r$. We denote by $\sigma_1 \geq \sigma_2 \geq \dots \geq \sigma_r$ the singular values of $A$. The perturbed matrix and its singular values set are denoted by $\tilde{A} = A + E$ and $\{ \tilde{\sigma}_j \}_{j=1}^r$, respectively. We write $\text{inf}_2(A)$ and $|A|_2$ to denote the smallest and largest singular values of any matrix $A$. The following classical result provides an estimated bound for the perturbed singular values~\citep{stewart1979note, stewart1998perturbation}.
\begin{thm} \label{thm:svd_perturb}
 Let $P$ be the orthogonal projection onto the column space of $A$. Let $P_{\perp} = I - P$. Then
 $$\tilde{\sigma}_j^2 = (\sigma_j + \gamma_j)^2 + \eta_j^2 \ , \quad j=1.\dots,r \ ,$$
\end{thm}
where $|\gamma_j| \leq |P \, E|_2$ and $\text{inf}_2(P_\perp E) \leq \eta_j \leq | P_\perp E |_2 \ .$

Following \cite{stewart1979note}, we make two observations with respect to Thm.~\ref{thm:svd_perturb}. First, if $\sigma_j \gg |E|_2$ then it dominates the bound and we have $\tilde{\sigma}_j \cong \sigma_j + \gamma_j$. Second and more important to our setting, when $\sigma_j$ is of order $|E|_2$, the term $\eta_j$ will tend to dominate. Indeed, in these cases the term $\eta_j$ \emph{increases} the singular value $\sigma_j$. We conclude that random perturbations to $A$ tend to increase its small singular values. In contrast, \code{\methodname} typically decreases the small singular values, while leaving the large $\sigma_j$ unchanged. Thus, \code{\methodname} is in effect a complementary approach to injecting additive noise, allowing a finer control over the resulting new samples. Finally, we note that for a certain choice of hyper-parameters, our approach can be viewed as injecting noise per the above analysis. For example, taking $\lambda \sim \text{Uniform}(1.0, \alpha)$ for $\alpha > 1.0$  will increase all the small singular values of $A$ by a factor of $\lambda \in [1.0, \alpha]$, where $\text{Uniform}$ is the random uniform distribution.

% For example, taking $\rho = 0.0$ and $\lambda \sim \text{Uniform}(1.0, \alpha)$ for $\alpha > 1.0$  will increase all the singular values of $A$ by a factor of $\lambda \in [1.0, \alpha]$, where $\text{Uniform}$ is the random uniform distribution.

% the corresponding VRM
\paragraph{\code{\methodname} as a Vicinal Risk Minimization (VRM).} Given a cost function $c: \mathcal{Y} \times \mathcal{Y} \rightarrow \mathbb{R}^{+}$, the learning problem aims at minimizing the expectation of the loss $c(f(x), y)$ over the distribution $\mathcal{P}(x, y), x\in\mathcal{X}, y\in\mathcal{Y}$. A fundamental challenge, shared by most real-world scenarios, is that the true distribution of the data is unfortunately \emph{unknown}. The alternative is to minimize over the empirical distribution of a train set $\{(x_i, y_i)\}_{i=1}^n$ given by
\begin{align*}
    \dd \mathcal{P}_\text{emp}(x,y) = \frac{1}{n} \sum_i \delta_{x_i}(x) \delta_{y_i}(y) \ .
\end{align*}
The resulting scheme is the common training procedure of modern neural networks, formally known as the Empirical Risk Minimization (ERM)~\citep{vapnik1991principles}. 

While $\mathcal{P}_\text{emp}$ provides a basic approximation of the true $\mathcal{P}$, it was suggested~\citep{chapelle2001vicinal} that other density estimates $\dd \mathcal{P}_\text{est}$ that take into account the \emph{vicinity} of $(x_i, y_i)$ should be considered. The recent \code{mixup} approach~\citep{guo2019mixup} exploits this idea by proposing a Vicinal Risk Minimization (VRM) procedure that is based on the vicinal distribution estimate $\frac{1}{n}\sum_{i,j} \delta_{\tilde{x}_{ij}(\lambda)}(x) \delta_{\tilde{y}_{ij}(\lambda)}(y)$, defined using convex combinations $\tilde{z}_{ij}(\lambda) = \lambda z_i + (1-\lambda) z_j$ for $z \in \{ x, y \}$ and $\lambda \sim \text{Beta}(\alpha, \alpha)$. In this context, the main difference between \code{\methodname} and \code{mixup} is in the definition of vicinity as we describe below.

We denote by $\mathcal{T}(x,y)$ the tangent plane of the data manifold $\mathcal{M}$ at the point $(x, y) \in \mathcal{M} \subset \mathcal{X} \times \mathcal{Y}$. Namely, $\mathcal{T}(x, y)$ is the linear approximation of $\mathcal{M}$ at $(x,y)$. For every pair $(x,y)$, we define a new density distribution $\mathcal{P}_\text{tan}$ which considers all pairs $(a,b)$ in the tangent plane of $(u,v) \in \mathcal{M}$. Formally,
\begin{equation*}
    \dd \mathcal{P}_\text{tan}(x, y) = \int_\mathcal{M} \int_{\mathcal{T}(u,v)} \delta_a(x) \delta_b(y) \dd ab \dd uv \ .
\end{equation*}
Then, \code{\methodname} approximates the latter expression by generating an estimate of the tangent plane $\mathcal{T}_\text{est}$ via \code{SVD}, yielding the following vicinal estimate
\begin{align*}
    & \dd \mathcal{P}_\text{est} (x,y) = \frac{1}{n} \sum_i \frac{1}{k_i} \sum_{j} \delta_{x_j} (x) \delta_{y_j} (y) \ , \\
    & (x_j, y_j) \in \mathcal{T}_\text{est}(x_i, y_i) \ , \; k_i = |\mathcal{T}_\text{est}(x_i, y_i)| \ .
\end{align*}

\section{Experiments}
\label{sec:results}

\begin{table*}[!t]
\centering
\caption{Comparison of out-of-distribution robustness problems. Bold values represent the best results and underlined values are second best. We report the average RMSE across domains and the ``worst within-domain'' RMSE over three different seeds. For the DTI and PovertyMap datasets, we report average $R$ and ``worst within-domain'' $R$. Full results with standard deviation can be found in App~\ref{app:res_std}.}
\label{tab:ood}
    \begin{tabular}{l|c|c|c|c|c|c|c}
    \toprule
    & RCF (RMSE) & \multicolumn{2}{|c}{Crimes (RMSE)} & \multicolumn{2}{|c}{DTI ($R$)} & \multicolumn{2}{|c}{PovertyMap ($R$)} \\ 
    \cline { 2 - 8 } & Avg. $\downarrow$  & Avg. $\downarrow$ & Worst $\downarrow$ & Avg. $\uparrow$ & Worst $\uparrow$ &Avg. $\uparrow$ & Worst $\uparrow$\\
    \midrule 
    ERM$^{\dagger}$ & 0.164 & 0.136 & 0.170 & 0.483  & 0.439 &0.80 & \underline{0.50}\\
    Mixup$^{\dagger}$ & 0.159 & 0.134 & 0.168 & 0.459 & 0.424 & \underline{0.81} & 0.46\\ 
    ManiMixup$^{\dagger}$ & $\underline{0.157}$ & $\mathbf{0.128}$ & $\mathbf{0.155}$ & 0.474 & 0.431 &- & -\\ 
    C-Mixup$^{\star}$ & $\mathbf{0.153}$ & $0.131$ & $0.166$ & 0.474 & \underline{0.441} & $0.803$ & \textbf{0.516} \\
    ADA$^{\star}$ & 0.171 & \underline{1.30} & \underline{0.156} & - & -& - & -\\
    \midrule
    \methodname & 0.171 & 0.132 & $0.164$ & $\underline{0.496}$ & $0.430$ & 0.776 & 0.492\\
    \methodname$_{\rho}$ & $0.159$ & \textbf{0.128} & 0.158  & $\mathbf{0.503}$ & $\mathbf{0.459}$& \textbf{0.832} & 0.482\\
    \midrule
    \midrule
    C-Mixup$^{\dagger}$ & 0.146 & 0.123 & 0.146 & 0.498 & 0.458 & 0.81 & 0.53 \\
    ADA$^{\dagger}$ & 0.175 & 0.130 & 0.156 & 0.493 & 0.448& 0.794 & 0.522\\
    \bottomrule
    \end{tabular}
\end{table*}

% datasets; citations; mention sklearn/UCI?

%1331
% \subsection{Computational resources}
% \label{app:speed}
% The computational complexity of \code{\methodname} is governed by \code{SVD} calculation which has a complexity of $\mathcal{O}(\min(qr^2,rq^2))$ for a $q\times r$ matrix. In Table~\ref{tab:speed_foma}, we compare the average epoch time across 50 epochs in seconds to provide an estimate for the empirical computational cost of \code{\methodname}. The results are obtained with a single \code{RTX3090} GPU. Note that the runtime of \code{\methodname} is very dependant on the batch size. Larger batch size results in less \code{SVD} computations, and depending on implementation, \code{SVD} of larger matrices can be evaluated faster.
%  % table for foma

% \begin{table*}[ht]
% \caption{Training times using \code{\methodname} (in seconds).}
% \label{tab:speed_foma}
% \centering
% \vskip 0.1in
%     \begin{tabular}{l|ccc}
%     \toprule 
%     Dataset & Airfoil & Electricity & DTI \\
%     \midrule 
%     ERM & 0.058 & 1.406& 33.374\\
    
%     \methodname - input & 0.148 & 3.101& 43.433 \\
%     \methodname - latent & 0.286 & 2.837& 51.399\\
%     \methodname - both & 0.321 & 4.376& 63.071 \\

%     \methodname $\rho$ - input & 0.119 & 2.806& 41.527 \\
%     \methodname $\rho$ - latent & 0.149  & 3.021& 43.985\\
%     \methodname $\rho$ - both & 0.262 & 4.315& 57.35 \\
%     \bottomrule
%     \end{tabular}
%     \vskip -0.1in
% \end{table*}

\subsection{In-Distribution Generalization}
\label{sec:id_gen}
In this section, we assess the effectiveness of \code{\methodname} and compare it to previous methods in terms of its ability to generalize within the given distribution. We utilize the datasets used in the study conducted by \citet{yao2022cmix} and closely replicate their experimental setup.

\paragraph{Datasets.} We use the following five datasets to evaluate the performance of in-distribution generalization. Two tabular datasets: Airfoil Self-Noise (Airfoil)~\citep{airfoil} and NO2~\citep{no2}. Airfoil includes the aerodynamic and acoustic test results of airfoil blade sections and NO2 predicts the level of air pollution at specific locations. Two time series datasets: Exchange-Rate and Electricity~\cite{lai2018modeling}, where Exchange-Rate provides a collection of daily exchange rates and Electricity is utilized for predicting the hourly electricity consumption. Finally, Echocardiogram Videos is designed for predicting the ejection fraction. It consists of a collection of videos that provide visual representations of the heart from various perspectives. See App.~\ref{app:datasets} for a detailed description of the datasets.

\paragraph{Experimental settings.} We conduct a comparative analysis between our approach, \code{\methodname}, and several existing strong baseline methods, namely Mixup~\citep{zhang2017mixup}, Manifold-Mixup~\citep{verma2019manifold}, C-Mixup~\citep{yao2022cmix}, Anchor Data Augmentation (ADA)~\citep{schneider2023anchor}, and classical expected risk minimization (ERM). Importantly, we denote by C-Mixup$^{\star}$ and ADA$^{\star}$ the results for these methods as reproduced in our environment, whereas C-Mixup$^{\dagger}$ and ADA$^{\dagger}$ represent the results as reported in~\cite{yao2022cmix, schneider2023anchor}, respectively. For a fair comparison, we compare our results with respect to the starred models, as we were unable to reproduce the reported results in the original papers; this observation also appeared in~\cite{schneider2023anchor} regarding C-Mixup. Further, to maintain consistency with the methodology outlined in~\citep{yao2022cmix}, we adopt the same model architectures, using a fully connected three-layer network for tabular datasets, an LST-Attn~\citep{lai2018modeling} for time series data, and EchoNet-Dynamic~\citep{ouyang2020video} for predicting the ejection fraction. We use Root Mean Square Error (RMSE) and Mean Averaged Percentage Error (MAPE) as evaluation metrics. Detailed experimental settings and hyper-parameters are provided in App.~\ref{app:id_hyper}.

\vspace{-3mm}

\paragraph{Results.} We report the in-distribution generalization results in Table~\ref{tab:id}. In all settings, lower numbers are better. Per column, we mark in bold the best available result, and we underline the second best error measure. Table~\ref{tab:id} shows that both variants of \code{\methodname} improve over standard training via ERM, often by large margins. Further, our approach achieves new state-of-the-art error measures in comparison to the other baseline approaches on all datasets and metrics, except for Electricity where we obtain on-par results to C-Mixup. In particular, \code{\methodname}$_\rho$ yields the best available results in most cases. Finally, we observe the most significant improvement occurs on two datasets: Airfoil and Exchange-Rate in which \code{\methodname} reduces the error of the state-of-the-art result by approximately $37\%$ and $38\%$, respectively.

\setlength{\tabcolsep}{2pt}
\begin{table*}[!t]
\centering
\caption{Ablation study of \code{\methodname} over modifying data at the input or latent levels, different loss scaling profiles $\mu(\lambda)$, and scaling down the small or large singular values.}
\label{tab:ablation}
\resizebox{\textwidth}{!}{
\begin{tabular*}{\textwidth}{@{\extracolsep{\stretch{1}}}ccc|cc|cc}
    \toprule
    & & & \multicolumn{2}{c|}{NO2} & \multicolumn{2}{c}{Electricity} \\ 
    mode & $\mu(\lambda)$ & scale & RMSE$\,\downarrow$ & MAPE$\,\downarrow$ & RMSE$\,\downarrow$ & MAPE$\,\downarrow$ \\
    \midrule
    input & $1$ & small & $\bm{0.52 \pm  0.01}$ & $13.23 \pm 0.09$ & $5.83e^{-2} \pm 1e^{-3}$ & $\bm{13.02} \pm \bm{0.07}$ \\
    input & $\lambda$ & small & $0.53 \pm 0.01$ &  $13.23 \pm 0.09$ & $\underline{5.83e^{-2} \pm 1e^{-3}}$ &  $13.02 \pm 0.37$ \\
    input & $\lambda^2$ & small & $0.54 \pm 0.01$ & $13.53 \pm 0.27$ & $\bm{5.79e^{-2} \pm 2e^{-4}}$ & $13.50 \pm 0.04$ \\
    \midrule
    input & 1 & large & $0.79 \pm 0.02$ & $19.22 \pm 0.45$ & $5.89e^{-2}$ $\pm$ $4e^{-4}$ & $13.89 \pm 0.39$ \\
    input & $\lambda$ & large & $0.76 \pm 0.02$ & $18.61 \pm 0.37$ & $5.86e^{-2} \pm 9e^{-4}$ & $\underline{13.15 \pm 0.05}$ \\
    input & $\lambda^2$ & large & $0.74 \pm 0.01$ & $18.01 \pm 0.28$ & $5.88e^{-2}$ $\pm$ $1e^{-3}$ & $13.23 \pm 0.13$ \\
    \midrule
    latent & $1$ & small & $0.53 \pm 0.01$ & $\underline{13.21 \pm 0.08}$ & $6.11e^{-2}$ $\pm$ $9e^{-4}$ & $15.58 \pm 0.38$ \\
    latent & $1$& large & $0.65 \pm 0.02$ & $16.35 \pm 0.56$ & $7.16e^{-2}$ $\pm$ $1e^{-3}$ & $18.62 \pm 0.70$ \\
    input + latent & $1$& small & $\underline{0.52 \pm \underline 0.01}$ & $\bm{13.19} \pm \bm{0.14}$ & $5.94e^{-2}$ $\pm$ $1e^{-3}$ & $14.78 \pm  0.48$ \\
    input + latent & $1$& large & $0.65 \pm 0.02$ & $21.25 \pm 0.48$ & $7.97e^{-2}$ $\pm$ $1e^{-4}$& $20.77 \pm 0.41$ \\
    \bottomrule
\end{tabular*} }
\end{table*}

\subsection{Out-of-Distribution Robustness}
\label{sec:ood_gen}

In this section, we assess the effectiveness of \code{\methodname} and compare it to previous methods on tasks involving out-of-distribution robustness. To this end, we consider the datasets used in the study~\citep{yao2022cmix}, and we closely replicate their experimental setup.

\vspace{-3mm}

\paragraph{Datasets.} We use the following four datasets to assess the performance of out-of-distribution robustness. \textbf{1)} RCFashionMNIST (RCF)~\citep{yao2022cmix}, which is a synthetic modification of Fashion-MNIST, modeling sub-population shifts, where the goal is to predict the angle of rotation for each object. \textbf{2)} Communities and Crime (Crime)~\citep{crime} is a tabular dataset that focuses on predicting the total number of violent crimes per 100K population, where the objective is to develop a model that can generalize to states that were not included in the training data. \textbf{3)} Drug-Target Interactions (DTI)~\citep{huang2021therapeutics} is aiming to predict out-of-distribution drug-target interactions where the year is the domain information. \textbf{4)} PovertyMap~\citep{koh2021wilds} is a satellite image regression dataset that has been created with the goal of estimating asset wealth in countries that were not included in the training set. For further details on the various datasets, see App.~\ref{app:datasets}.

\paragraph{Experimental settings.} Similarly to Sec.~\ref{sec:id_gen}, we compare \code{\methodname} to the same baseline approaches. In terms of metrics, we report the RMSE (lower is better) for RCF-MNIST and Crimes. For PovertyMap and DTI, we use $R$ (higher is better) as the evaluation metric, originally proposed in the respective papers~\cite{koh2021wilds, huang2021therapeutics}. Following~\citet{yao2022cmix}, we trained ResNet-18 for RCF-MNIST and PovertyMap datasets, three-layer fully connected networks for Crimes, and DeepDTA~\cite{ozturk2018deepdta} for DTI. More details regarding hyper-parameters and the experimental setup appear in App.~\ref{app:id_hyper}.

% add results on poverty
\paragraph{Results.} We report both the average and worst-domain performance metrics for out-of-distribution tasks in Table~\ref{tab:ood}. Dashed cells represent cases where results are not available. While the results for our approach are somewhat more mixed in comparison to the in-distribution challenge, we still find \code{\methodname} to be highly effective. In particular, \code{\methodname} improved over ERM in almost all cases, with the exception of RCF. We find our technique to obtain comparable results for RCF and Crimes with respect to the best baselines. Notably, \code{\methodname} presents a noticeable gap in DTI and PovertyMap in comparison to SOTA approaches such as C-Mixup and ADA. Finally, similarly to the in-distribution setting, \code{\methodname}$_\rho$ achieves better results in comparison to \code{\methodname}.

\subsection{Ablation study}
\label{subsec:abl}

\code{\methodname} is a data augmentation method that scales down singular values of the data. However, there are several design choices to make. For example, we can choose to apply \code{\methodname} on the input data, on the learned representations, or apply it on both in succession. Another decision is what singular values we should scale down: the smaller ones, capturing less explained variance of the data, or the larger ones, which may create samples that are further away from the data manifold and thus expand the underlying distribution. Another choice to make is how to scale the singular values and potentially the loss function. The values $\lambda \sim \text{Beta}(\alpha, \alpha)$ by which we scale the singular values vary between batches, the higher the value of $\lambda$, the more noise is removed. By scaling the loss function differently for each batch, we can change the update rule of the optimizer, potentially leading to an improved behavior of \code{\methodname}.

We detail in Table~\ref{tab:ablation} the ablation results we obtained for NO2 and Electricity datasets while exploring the parameter spaces of the above design choices. Overall, applying our technique at the input level, without scaling the loss, and scaling the small singular values seems to consistently yield good results across datasets and tasks (see also Tables~\ref{tab:hyp_foma}, \ref{tab:hyp_foma_rho}). More specifically, we find in Table~\ref{tab:ablation} that for the datasets NO2 and Electricity, scaling down the small singular values is preferred to scaling the larger ones and more generally, this observation holds for all datasets (see Table~\ref{tab:hyp_foma}). On the other hand, when using \code{\methodname}$_{\rho}$, allowing more freedom in selecting what singular values to scale, some datasets achieve better results when scaling the \emph{larger} singular values, e.g., Airfoil and Exchange-rate. Furthermore, we have two observations regarding the loss scaling profiles $\mu (\lambda)$: 1) the effect of decreasing $\mu$ is inversely proportional between scaling small singular values and large singular values. For instance, when scaling the small singular values of NO2, smaller $\mu$ corresponds with better performance, whereas when scaling the large singular values, larger $\mu$ corresponds to better performance. 2) The effect of $\mu$ is inversely proportional between the datasets, decreasing $\mu$ improves the performance on Electricity while achieving worse performance on NO2. Finally, we note that employing \code{\methodname} in the latent space or both in the input and latent spaces yields inconsistent results across datasets.

\section{Discussion}
\label{sec:discussion}

We have proposed \code{\methodname}, a data-driven method for data augmentation of regression tasks. We showed that \code{\methodname} supports the network tendency of representing dominant components of its input signals by creating virtual examples sampled from the tangent planes of the original train set. Implementing \code{\methodname} is straightforward, and it is fully differentiable. Throughout an extensive evaluation, we have shown that \code{\methodname} improves the generalization error of neural models on regression benchmarks including in-distribution generalization and out-of-distribution tasks. We also ablated our model with respect to several design choices. Generally, we find our method to obtain highly competitive results and often surpass state-of-the-art approaches.

% limitations: choice of parameters, computational complexity, dependence on tangent space
We inspected the effect of the hyper-parameters $\alpha$, $\rho$ and whether to scale the small or large singular values. For \code{\methodname}, we observe a relatively consistent performance across tasks, whereas \code{\methodname}$_\rho$ presents mixed results (see Tabs.~\ref{tab:hyp_foma}, \ref{tab:hyp_foma_rho}). Given that \code{\methodname} and \code{\methodname}$_\rho$ perform similarly, we can not derive a specific guideline for choosing these hyper-parameters. Another limitation is that the time complexity of \code{\methodname} is governed by the \code{SVD} calculation, which may be restrictive for large train batches. Finally, we mention that if $k=n_0+m$, i.e., the tangent space dimension equals that of the data rank, then our approach can be only applied to the dominant singular values.

There are several exciting avenues for future exploration. First, is there a fundamental link between the vicinal distribution employed and the learned representation? While several existing works suggest that \emph{linearity} yields better models~\cite{azencot2020forecasting, berman2023multifactor, zeng2023transformers}, the model dependency on the specific definition of vicinity is still not well understood. Second, can similar methods to ours be useful in classification tasks? The adaptation of \code{\methodname} to classification is straightforward, however, several design choices which were tuned for regression may require change in a classification setting. In particular, the computational demands of SVD-based data augmentation are higher in comparison to mixup schemes. Improving these aspects by e.g., approximating the tangent space of the manifold may be highly impactful in regression as well as classification tasks. Another interesting avenue to explore is the relation between \code{FOMA} and generative modeling~\cite{naiman2024generative}. We plan to explore these questions in future work.

% \section{Acks}
\clearpage
\section*{Acknowledgements}
\label{Sec:acknowledgement}
This research was partially supported by the Lynn and William Frankel Center of the Computer Science Department, Ben-Gurion University of the Negev, an ISF grant 668/21, an ISF equipment grant, and by the Israeli Council for Higher Education (CHE) via the Data Science Research Center, Ben-Gurion University of the Negev, Israel.
\section*{Impact Statement}

This paper presents work whose goal is to advance the field of Machine Learning. There are many potential societal consequences of our work, none which we feel must be specifically highlighted here.

\clearpage

\bibliographystyle{icml2024}
\bibliography{main}

%%%%%%%%%%%%%%%%%%%%%%%%%%%%%%%%%%%%%%%%%%%%%%%%%%%%%%%%%%%%%%%%%%%%%%%%%%%%%%%
%%%%%%%%%%%%%%%%%%%%%%%%%%%%%%%%%%%%%%%%%%%%%%%%%%%%%%%%%%%%%%%%%%%%%%%%%%%%%%%
% APPENDIX
%%%%%%%%%%%%%%%%%%%%%%%%%%%%%%%%%%%%%%%%%%%%%%%%%%%%%%%%%%%%%%%%%%%%%%%%%%%%%%%
%%%%%%%%%%%%%%%%%%%%%%%%%%%%%%%%%%%%%%%%%%%%%%%%%%%%%%%%%%%%%%%%%%%%%%%%%%%%%%%
\clearpage
\appendix
\onecolumn

\section{Method overview}

\label{app:overview}

In this section we provide details for different methods used to evaluate the intrinsic dimension and linear dimension which are used for \code{FOMA} and \code{FOMA$\rho$} respectively.

\subsection{Intrinsic dimension}
\label{app:id}
To estimate the intrinsic dimension of data, we use the TwoNN~\citep{facco2017estimating} ID estimator. The ID-estimator relies on the distances to only the two closest neighbors of each point, minimizing the influence of inconsistencies in the dataset during estimation.
\\
Let $X=\{x_{1},x_{2}, \cdots , x_{N}\}$ be a set of points sampled uniformly on a manifold with intrinsic dimension $d$. For each point $x_{i}$, we calculate the two shortest distances $r_{1}, r_{2}$ from other elements in $X \setminus \{ x_{i} \}$ and determine the ratio $\mu_{i}=\frac{r_{2}}{r_{1} }$. It has been proven that $\mu_{i}, 1 \leq i \leq N$ are distributed according to a Pareto distribution with parameter $d + 1$ on the interval $[1,\infty)$, specifically $f\left(\mu_i \mid d\right)=d \mu_i^{-(d+1)}$. While $d$ can be estimated by maximizing the likelihood:
\begin{equation}
    \label{eqn:twonn_like}
    P(\mu_{1}, \mu_{2}, \cdots \mu_{N} \mid d)=d^N \prod_{i=1}^N \mu_i^{-(d+1)} \ .
\end{equation}
\\
We adopt the approach suggested by~\citet{facco2017estimating} using the cumulative distribution $F(\mu)=1-\mu^{-d}$. The method involves estimating the parameter $d$ through linear regression on the empirical estimate of $F(\mu)$. To do this, we arrange the values of $\mu$ in ascending order and define $F^{emp}(\mu_{i}) \approx \frac{i}{N}$. A linear regression is then performed on the set of points $\{(\log \mu_i,-\log (1-F_i^{emp}))\}_{i=1}^N$ where the slope of the fitted line is the estimated ID.

\subsection{Linear dimension}
\label{app:linear_id}
A common method for estimating the linear dimension of data is to perform principal component analysis (PCA) or SVD and count the number of components that should be included to describe some percentage of the variance in the data, usually above 90\%. More formally:
\begin{align*}
    k = \arg\max_{\tilde{k}} \sum_{j=1}^{\tilde{k}} \sigma_j / \sum_j \sigma_j \leq \rho \ .
\end{align*}

\section{Sequential models capture dominant components of data better}
\label{app:models_lamv}

\begin{figure*}[t!]
  \centering
  \begin{overpic}[width=1\linewidth]{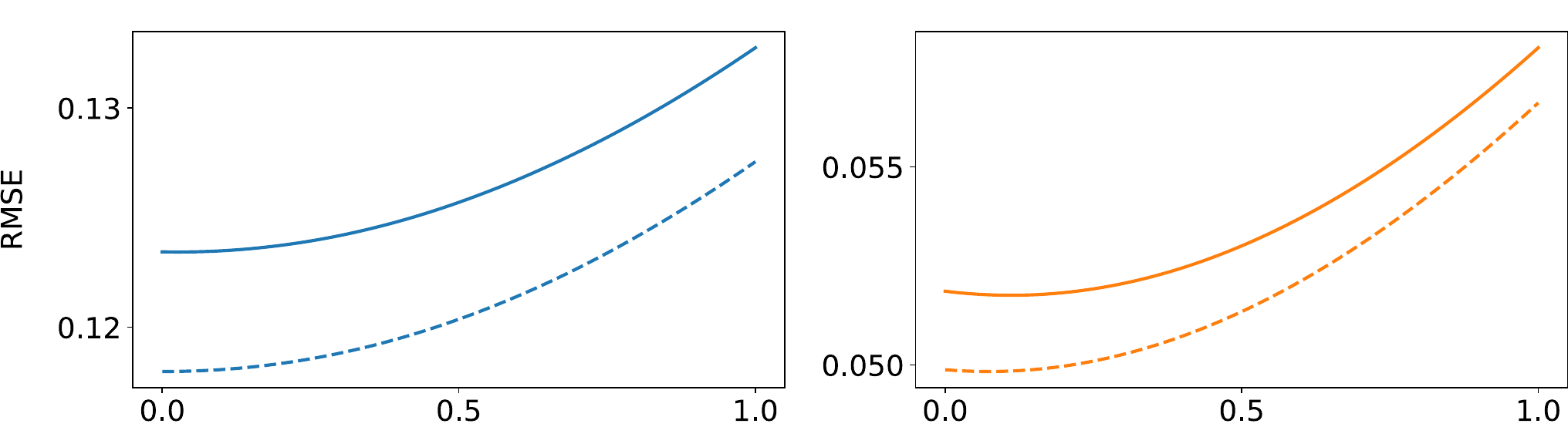}
      \put(26, 26){Crimes} \put(75, 26){Electricity}
  \end{overpic}
  \caption{Evaluating a non-augmented model (solid lines) and a model trained with \code{\methodname} (dashed lines) on train data whose small singular values are scaled down for different values of $\lambda$ (see also Fig.~\ref{fig:cz_analysis}, left). Communities and crime dataset trained on a three-layer full connected network (left). Electricity trained on LST-Attn~\citep{lai2018modeling} (right).}  
  \label{fig:models_lamv_app}
\end{figure*}

Following the discussion in Sec.~\ref{sec:method}, we verify empirically that neural networks model the dominant parts of their data better. We repeat the experiment in Fig.~\ref{fig:cz_analysis} (left) using additional two datasets which are trained on different architectures. For evaluation, we use the dataset whose singular values are modified using varying values of $\lambda$. The results are presented in Fig.~\ref{fig:models_lamv_app}, where solid lines represent the results of the non-regularized model, and dashed lines are associated with models trained with \code{\methodname}. Remarkably, we observe a similar qualitative behavior as we reported in Sec.~\ref{sec:method}. In particular, the highest MSE values are obtained for both the baseline and regularized models for $\lambda=1$, i.e., when the data is unchanged. Further, the model attain improved error measures as $\lambda$ decreases, where the error profile is similar for the baseline and regularized models. Based on these results, we deduce that sequential models prefer to represent and compute the dominant components of data, reinforcing our choice for supplying such data to the network during training.

\section{Dataset Description}
\label{app:datasets}

In this section, we provide detailed descriptions of datasets used in the experiments in this work.

\paragraph{Airfoil Self-Noise~\cite{airfoil}.} The dataset comprises aerodynamic and acoustic test findings for various sizes of NACA 0012 airfoils, obtained at different wind tunnel speeds and angles of attack. Each input instance consists of five features: frequency, angle of attack, chord length, free-stream velocity, and suction side displacement thickness. The label represents the one-dimensional scaled sound pressure level. To normalize the input features, min-max normalization is applied. As per reference~\cite{hwang2021regmix}, the training, validation, and test sets consist of 1003, 300, and 200 examples, respectively.

\paragraph{NO2~\cite{no2}.} The NO2 emission dataset originates from a study examining the relationship between air pollution near a road and traffic volume along with meteorological variables. Each input comprises seven features, including the logarithm of the number of cars per hour, temperature 2 meters above ground, wind speed, temperature difference between 25 and 2 meters above ground, wind direction, hour of the day, and the day number since October 1st, 2001. The hourly values of the logarithm of NO2 concentration, measured at Alnabru in Oslo between October 2001 and August 2003, serve as the response variable or label. As per reference~\cite{hwang2021regmix}, there are 200 examples in the training set, 200 in the validation set, and 100 in the test set.

\paragraph{Exchange-Rate~\cite{lai2018modeling}.} The exchange-rate dataset comprises a time-series collection of daily exchange rates for eight countries: Australia, Britain, Canada, Switzerland, China, Japan, New Zealand, and Singapore, spanning from 1990 to 2016. The total length of the time series is 7,588, with a daily sampling frequency. A sliding window size of 168 days is applied. The input dimension is 168 × 8, and the label dimension is 1 × 8 data points. Following the methodology outlined in~\cite{lai2018modeling}, the dataset has been divided into training (60\%), validation (20\%), and test (20\%) sets in chronological order.

\paragraph{Electricity~\cite{airfoil}.} This dataset is a time-series collection obtained from 321 clients, representing electricity consumption in kWh recorded every 15 minutes from 2012 to 2014. The total length of the time series is 26,304, sampled hourly. As with the Exchange-Rate data, a window size of 168 is utilized, resulting in an input dimension of 168 × 321 and a corresponding label dimension of 1 × 321. Similar to the methodology described in Lai et al.~\cite{lai2018modeling}, the dataset is divided accordingly.

\paragraph{Echo~\cite{ouyang2020video}.} The Echocardiogram Videos dataset comprises 10,030 labeled apical-4-chamber echocardiogram videos captured from various perspectives, accompanied by expert annotations for studying cardiac motion and chamber sizes. These videos were sourced from individuals undergoing imaging at Stanford University Hospital between 2016 and 2018. To delineate the left ventricle area, initial preprocessing involves frame-by-frame semantic segmentation of the videos. This preprocessing method generates video clips containing 32 frames of 112 × 112 RGB images, which serve as input for predicting ejection fraction. The dataset is partitioned into training, validation, and test sets, with sizes of 7,460, 1,288, and 1,276, respectively.

\paragraph{RCF.} 
RCF-MNIST, where "RCF" stands for "Rotated-Colored-Fashion", is a dataset constructed with specific color and rotation attributes. In this dataset, the normalized RGB vector for red and blue is $[1, 0, 0]$ and $[0, 0, 1]$ respectively, and the normalized rotation angle (i.e., label) for each image is denoted as $g$, where $g \in [0, 1]$. During the construction of the training set, 80\% of the images are colored using the RGB value 
$[g, 0, 1-g]$, while the remaining 20\% are colored with $[1-g, 0, g]$. Consequently, there is a strong spurious correlation between color information and labels within the training set. To simulate distribution shift in the test set, this spurious correlation is reversed, with 80\% of the images colored using RGB values $[1-g, 0, g]$, and the remaining 20\% with $[g, 0, 1-g]]$. The impact of this spurious correlation on performance is evaluated by comparing the performance of the same test set with or without distribution shift. The results, presented in Table 11, demonstrate that the subpopulation shift induced by the spurious correlation indeed affects performance negatively, as anticipated.

\paragraph{PovertyMap.} This dataset is part of the WILDS benchmark~\cite{koh2021wilds}, comprising satellite images sourced from 23 African countries, which are utilized for predicting the village-level real-valued asset wealth index. Each input consists of a 224 × 224 multispectral LandSat satellite image with 8 channels, while the corresponding label represents the real-valued asset wealth index. The domains of the images encompass information regarding the country, urban, and rural areas. The dataset is divided into 5 distinct cross-validation folds, with all countries in these splits being disjoint to facilitate the out-of-distribution setting. All experimental configurations adhere to the methodology outlined by Koh et al.~\cite{koh2021wilds}.

\paragraph{Crime~\cite{crime}.} The Communities And Crimes dataset is a tabular compilation that integrates socio-economic data from the 1990 US Census, law enforcement data from the 1990 US LEMAS survey, and crime data from the 1995 FBI UCR. It encompasses 122 attributes believed to have some plausible connection to crime, such as median family income and the percentage of officers assigned to drug units. The target attribute for prediction is per capita violent crimes, which includes offenses such as murder, rape, robbery, assault, among others. To prepare the data, all numeric features are normalized using the decimal range 0.00 to 1.00 through an equal-interval binning method, and missing values are imputed with the average values of the corresponding attributes. Domain information is denoted by state identifications, resulting in a total of 46 domains. The dataset is partitioned into training, validation, and test sets containing 1,390, 231, and 373 instances, respectively. These sets consist of 31, 6, and 9 disjoint domains, respectively.

% % SkillCraft. SkillCraft is a UCI tabular dataset [6] originated from a study that used video game telemetry data from real-time strategy (RTS) games to explore the development of expertise. Input x contains 17 player-related parameters in the game, such as the Cognition-Action-cycle variables and the Hotkey Usage variables. And the action latency in the game was considered as the label y. Missing data are filled by mean padding on each attribute. We use "League Index", which correspond to different levels of competitors, to be the identifier of domain. The dataset is split into training, validation and test sets with size 1878, 806, 711 and disjoint domain number 4, 1, 3, respectively.

\paragraph{DTI~\cite{huang2021therapeutics}.} The Drug-target Interactions dataset is designed to forecast the binding activity score between each small molecule and its corresponding target protein. Input features encompass information on both the drug and target protein, represented by one-hot vectors, while the output label denotes the binding activity score. The training and validation sets are curated from the years 2013 to 2018, while the test set spans the years 2019 to 2020. The "Year" attribute serves as the domain information.
\newpage
\section{Additional Experiments}

\label{app:add_exps}

\subsection{Comparison with additive noise}
\label{app:noise}

In Section.~\ref{sec:analysis} we show a relation of our method to additive noise. For completeness, we add a supplementary experiment that compares our method with additive noise as shown in Table.~\ref{tab:app_noise} and Table.~\ref{tab:noise_ood}. The experiment was conducted as follows: Given a sample $x_i \in \mathbb{R}^n$ and $y_{i} \in \mathbb{R}^m$
we sample  $\epsilon_x \sim \mathcal{N} (0,\sigma \times I_{n}),\ \epsilon_y \sim \mathcal{N} (0,\sigma \times I_{m})$ and create the noised sample $(\Tilde{x_i} = x_i+\epsilon_x, \Tilde{y_i = y_i +\epsilon_y})$. Where $\sigma$ is selected from $\{0.1, 0.01,0.001,0.0001 \}$ and the best model is chosen according to the performance on the validation set.

\begin{table*}[h]
\centering

\caption{Comparison of in-distribution generalization tasks with additive noise.}
\label{tab:app_noise}

\vskip 0.1in

% \resizebox{\textwidth}{!}{
\begin{tabular}{l|c|c|c|c|c|c|c|c|c|c}
\hline & \multicolumn{2}{|c}{ Airfoil } & \multicolumn{2}{|c}{ NO2 } & \multicolumn{2}{|c}{ Exchange-Rate } & \multicolumn{2}{|c}{ Electricity } \\
\cline { 2 - 9 }& RMSE $\downarrow$ & MAPE (\%)$\downarrow$ & RMSE $\downarrow$ & MAPE (\%)$\downarrow$ & RMSE $\downarrow$ & MAPE (\%)$\downarrow$ & RMSE $\downarrow$ & MAPE (\%)$\downarrow$ \\
\hline 
ERM$^{\dagger}$ & 2.901 & 1.753 & 0.537 &  13.615 & 0.0236 & 2.423 & 0.0581  & 13.861\\
ERM+noise & 3.172 & 1.864 & 0.524 & 13.326 & 0.016 & 1.568 & 0.600 & 14.068 \\

\hline
\methodname & $\underline{1.646 }$ & $\underline{0.963 }$ & $0.521 $ & $13.23$ & $\underline{0.013}$ & $\underline{1.280}$ & $\underline{0.058}$ & 14.614  \\
\methodname$_{\rho}$ & $\mathbf{1.471}$ & $\mathbf{0.816}$ & $\mathbf{0.512 }$ & $\mathbf{12.894}$ & $\mathbf{0.013 }$ & $\mathbf{1.262}$ & $0.059$ & \underline{13.437}  \\
\bottomrule
\end{tabular}

\end{table*}

\begin{table*}[h]
\centering
\caption{Comparison of out-of-distribution robustness problems with additive noise.}
\label{tab:noise_ood}
\vskip 0.1in

    \begin{tabular}{l|c|c|c|c|c|c|c|}
    % \toprule
    & RCF (RMSE) & \multicolumn{2}{|c}{Crimes (RMSE)} & \multicolumn{2}{|c}{DTI ($R$)} \\ 
    \cline { 2 - 6 } & Avg. $\downarrow$  & Avg. $\downarrow$ & Worst $\downarrow$ & Avg. $\uparrow$ & Worst $\uparrow$ \\
    \midrule 
    ERM$^{\dagger}$ & 0.164 & 0.136 & 0.170 & 0.483  & 0.439 \\
    ERM + noise & 0.180 & 0.136 & 0.166 & 0.492 & 0.442 \\ 
    \midrule
    \methodname & 0.171 & 0.132 & $0.164$ & $\underline{0.496}$ & $0.430$ \\
    \methodname$_{\rho}$ & $0.159$ & \textbf{0.128} & 0.158  & $\mathbf{0.503}$ & $\mathbf{0.459}$\\
    \bottomrule
    \end{tabular}
\end{table*}

\section{Hyperparameters}
\label{app:id_hyper}

We list the hyperparameters for every dataset in Table~\ref{tab:hyp_foma} and Table~\ref{tab:hyp_foma_rho} for the methods \code{\methodname} and \code{\methodname}$\rho$, respectively. In our main results, we apply our method on the input space or on the latent space or both and report the one with best performance. All hyperparameters are selected by cross-validation, evaluated on the validation set. Some of the hyperparameters such as architecture and optimizer are not included in the table since they were not changed and were used as they appear in previous works~\cite{yao2022cmix}.

 % table for foma
\begin{table*}[h]

\caption{Hyperparameter choices for the experiments using \code{\methodname}.}
\label{tab:hyp_foma}
\centering
\vskip 0.1in
    \begin{tabular}{l|ccccc|cccc}
    \toprule 
    Dataset & Airfoil & NO2 & Exchange-Rate & Electricity & Echo & RCF & Crimes & PovertyMap & DTI \\
    \midrule 
    Learning rate & $5 \mathrm{e}^{-4}$ &  $1 \mathrm{e}^{-3}$ & $1 \mathrm{e}^{-2}$ & $5 \mathrm{e}^{-4}$ &$1 \mathrm{e}^{-4}$ &$1 \mathrm{e}^{-4}$ &  $5 \mathrm{e}^{-3}$ & $1 \mathrm{e}^{-3}$ &  $5 \mathrm{e}^{-4}$\\
    
    Batch size & 32 & 64 & 16 & 128 & 10 &128 & 8 & 32 & 64 \\
    Input/Latent & latent & both &  input & both & latent &latent & input & latent &latent\\
    Epochs & 200 & 100 & 200 & 150 & 20 &250 & 200 & 50 & 200\\
    Singular values & small & small &  small & small & small &small &small & small & large\\
    $\alpha$  & 0.8 & 0.9 & 0.2 & 1.1 & 1.1 &1 & 0.6 & 1.5 & 0.5\\
    % Batch method & 1 & 0 &1 & 2 & 0&0 & 2 & 0\\

    %Sampling method & kde & - & kde & -&-  \\
    % Id & input & input & - & input & &batch & input\\
    \bottomrule
    \end{tabular}
    \vskip -0.1in
\end{table*}

\clearpage

% % table for foma rho
\begin{table*}[!t]

\caption{Hyperparameter choices for the experiments using \code{\methodname}$\rho$.}
\label{tab:hyp_foma_rho}
\centering
\vskip 0.1in
    \begin{tabular}{l|ccccc|cccc}
    \toprule
    Dataset & Airfoil & NO2 & Exchange-Rate & Electricity & Echo & RCF & Crimes & PovertyMap & DTI \\
    \midrule
    Learning rate & $5 \mathrm{e}^{-4}$ & $1 \mathrm{e}^{-3}$& $1 \mathrm{e}^{-3}$ & $1 \mathrm{e}^{-4}$  & $1 \mathrm{e}^{-4}$ &$1 \mathrm{e}^{-4}$  & $1 \mathrm{e}^{-3}$ & $5 \mathrm{e}^{-3}$ & $1 \mathrm{e}^{-2}$\\
    
    Batch size & 128 & 8 & 8 & 8 & 10 & 8 & 64 & 32 & 64\\
    Input/Latent  & input & input & input & input & latent & latent & both & latent & latent\\
    Epochs & 100 & 250 & 150 & 200 & 20 & 250 & 100& 50 & 250\\
    Singular values & large & small & large & large & small & small & large & small & large \\
    $\alpha$ & 1.4 & 0.3 & 1 & 0.7 & 1.1 & 1.5 & 0.6 & 1 & 0.6\\
    % Batch method & 1 & 1 & 0 & 0 & 2 & 2 & 1 & 0\\

    $\rho$ & 0.975 & 0.95 & 0.8 & 0.875 & 0.85 & 0.95 & 0.875& 0.875&0.825\\
    %Sampling method & knn & kde & & & kde & kde\\
    \bottomrule
    \end{tabular}
    \vskip -0.1in
\end{table*}

\section{Results with Standard Deviation}
\label{app:res_std}
In Tables~\ref{tab:id_std}, \ref{tab:ood_std}, we report the full results of in-distribution generalization and out-of-distribution robustness respectively.

\begin{table*}[ht!]
\centering
\caption{Full results of in-distribution generalization. We compute the mean and standard deviation
for results of three seeds.}
\label{tab:id_std}

\vskip 0.1in
\begin{tabular}{l|c|c|c|c|c|c|}
\hline & \multicolumn{2}{|c}{ Airfoil } & \multicolumn{2}{|c}{ NO2 } & \multicolumn{2}{|c}{ Exchange-Rate }\\
\cline { 2 - 7 }& RMSE $\downarrow$ & MAPE (\%)$\downarrow$ & RMSE $\downarrow$ & MAPE (\%)$\downarrow$ & RMSE $\downarrow$ & MAPE (\%)$\downarrow$ \\
\hline
ERM & $\underline{2.901 \pm 0.067}$ & $1.753 \pm 0.078$ & $0.537 \pm 0.005$ &  $13.615 \pm 0.165$ & $0.023 \pm 0.003$ & $2.423 \pm 0.365$ \\
Mixup & $3.730 \pm 0.190$ & $2.327 \pm 0.159$ &  $0.528 \pm 0.005$ & $13.534 \pm 0.125$ & $0.023 \pm 0.002$ & $2.441 \pm 0.286$ \\
Mani Mixup & $3.063 \pm 0.113$ & $1.842 \pm 0.114$ & $0.522 \pm 0.008$ & $13.357 \pm 0.214$ & $0.024 \pm 0.004$ & $2.475 \pm 0.346$\\
C-Mixup$^{\star}$ & $2.739 \pm 0.06$ & $1.640 \pm 0.069$ & $0.516 \pm 0.01$ & $13.069 \pm 0.294$ & $0.024 \pm 0.005$ & $2.455 \pm 0.629$\\
ADA$^{\star}$ & $2.357 \pm 0.118$ & $1.377 \pm 0.064$ & $\underline{0.515 \pm 0.006}$ & $13.128 \pm 0.12$ & $0.021 \pm 0.006$ & $2.250 \pm 0.781$\\
\hline
\methodname & $\underline{1.646 \pm 0.103}$ & $\underline{0.963 \pm 0.056}$ & $0.521 \pm 0.013$ & $13.23 \pm 0.289$ & $\underline{0.013 \pm 0.000}$ & $\underline{1.280 \pm 0.037}$ \\
\methodname$_{\rho}$ & $\mathbf{1.471 \pm 0.047}$ & $\mathbf{0.816 \pm 0.008}$ & $\mathbf{0.512 \pm 0.008}$ & $\mathbf{12.894 \pm 0.217}$ & $\mathbf{0.013 \pm 0.000}$ & $\mathbf{1.262 \pm 0.037}$ \\
\hline
\hline
C-Mixup$^{\dagger}$ & $2.717 \pm 0.067$ & $1.610 \pm 0.085$ & $0.509 \pm 0.006$ & $12.998 \pm 0.271$ & $0.0203 \pm 0.001$ & $2.041 \pm 0.134$ \\
ADA$^{\dagger}$ & $2.360 \pm 0.133$ & $1.373 \pm 0.056$ & $0.514 \pm 0.007$ & $13.127 \pm 0.146$ & $0.020 \pm 0.006$ & $2.115 \pm 0.689$ \\
\hline
\cline { 2 - 5 }
\centering
& \multicolumn{2}{|c}{ Electricity } & \multicolumn{2}{|c|}{ Echo }\\
\cline { 2 - 5 }& RMSE $\downarrow$ & MAPE (\%)$\downarrow$ & RMSE $\downarrow$ & MAPE (\%)$\downarrow$ \\
\cline { 1 - 5 }
ERM & $0.058 \pm 0.001$ & $13.861 \pm 0.152$ &  $5.402 \pm 0.024$ & $8.700 \pm 0.015$  \\
Mixup & $0.058 \pm 0.000$ & $14.306 \pm 0.048$ & $5.393 \pm 0.040$ & $8.838 \pm 0.108$ \\
Mani Mixup & $0.058 \pm 0.000$ & $14.556 \pm 0.057$ & $5.482 \pm 0.066$ & $8.955 \pm 0.082$\\
C-Mixup$^{\star}$ & $\mathbf{0.057 \pm 0.000}$ & $13.471 \pm 0.15$ &$5.483 \pm 0.097$ & $9.121 \pm 0.208$\\
ADA$^{\star}$ &$0.059 \pm 0.001$ & $13.578 \pm 0.146$ & - & - \\
\cline { 2 - 5 }
\methodname & $\underline{0.058 \pm 0.000}$ & $\underline{14.653 \pm 0.166}$ & $\mathbf{5.215 \pm 0.061}$ & $8.331 \pm 0.088$\\
\methodname$_{\rho}$ & $0.059 \pm 0.000$ & $\mathbf{13.437 \pm 0.26}$ & ${5.476 \pm 0.01}$ & $\mathbf{8.742 \pm 0.091}$  \\
\cline { 2 - 5 }
\cline { 2 - 5 }
C-Mixup$^{\dagger}$ & $0.057 \pm 0.001$ & $13.372 \pm 0.106$ & $5.177 \pm 0.036$ & $8.435 \pm 0.089$ \\
ADA$^{\dagger}$ & $0.058 \pm 0.001$ & $13.464 \pm 0.296$ & -& -\\
\cline { 2 - 5 }
\end{tabular}
\vskip -0.1in
\end{table*}

\begin{table*}[!t]
\centering
\caption{Full results of out-of-distribution generalization. We compute the mean and standard deviation
for results of three seeds.}
\vskip 0.1in
\label{tab:ood_std}
\resizebox{\textwidth}{!}{
    \begin{tabular}{l|c|c|c|c|c|c|c}
    \toprule
    & RCF (RMSE) & \multicolumn{2}{|c}{Crimes (RMSE)} & \multicolumn{2}{|c}{DTI ($R$)} & \multicolumn{2}{|c}{PovertyMap  ($R$)} \\ 
    \cline { 2 - 8 } & Avg. $\downarrow$  & Avg. $\downarrow$ & Worst $\downarrow$ & Avg. $\uparrow$ & Worst $\uparrow$ &Avg. $\uparrow$ & Worst $\uparrow$\\
    \midrule 
    ERM & $0.162 \pm 0.003$ & $0.134 \pm 0.003$ & $0.173 \pm 0.009$ & $0.464 \pm 0.014$ & $0.429 \pm 0.004$ & $0.80 \pm 0.04$ & $\underline{0.50 \pm 0.07}$\\
    Mixup &$0.176 \pm 0.003$ & $0.128 \pm 0.002$ & $\mathbf{0.154 \pm 0.001}$ & $0.465 \pm 0.004$ & $0.437 \pm 0.016$ & $\underline{0.81 \pm 0.04}$ & $0.46 \pm 0.03$\\ 
    ManiMixup & $0.157 \pm 0.020$ & $0.128 \pm 0.003$ & $\underline{0.155 \pm 0.009}$ &$0.474 \pm 0.004$ & $0.431 \pm 0.009$&- & -\\ 
    C-Mixup$^{\star}$ & $\mathbf{0.153 \pm 0.004}$ & $\underline{0.130 \pm 0.003}$ & $0.161 \pm 0.01$ & $0.475 \pm 0.013$ & $0.440 \pm 0.016$ & $0.804 \pm 0.03$ & $\mathbf{0.517 \pm 0.06}$\\
    ADA$^{\star}$ & $0.171 \pm 0.009$ & $0.130 \pm 0.003$  & $0.156 \pm 0.006$ & - & -& - & -\\
    \midrule
    \methodname & $0.171 \pm 0.015$ & $0.132 \pm 0.002$ & $0.164 \pm 0.002$ & $\underline{0.492 \pm 0.003}$ & $\underline{0.442 \pm 0.019}$ & $0.776 \pm 0.03$ & $0.492 \pm 0.05$\\
    \methodname$_{\rho}$ &$\underline{0.159 \pm 0.01}$& $\mathbf{0.128 \pm 0.004}$ & $0.158 \pm 0.002$  & $\mathbf{0.503 \pm 0.008}$ & $\mathbf{0.459 \pm 0.01}$& $\mathbf{0.832 \pm 0.04}$ & $0.482 \pm 0.06$\\
    \midrule
    \midrule
    C-Mixup$^{\dagger}$ & $0.146 \pm 0.005$ & $0.123 \pm 0.000$ & $0.146 \pm 0.002$ & $0.498 \pm 0.008$ & $0.458 \pm 0.004$ & $0.81 \pm 0.03$ & $0.53 \pm 0.07$ \\
    ADA$^{\dagger}$ & $ 0.162 \pm 0.014 $ & $ 0.129 \pm 0.003 $ & $ 0.155 \pm 0.006 $ & $ 0.492 \pm 0.009 $ & $ 0.448 \pm 0.009 $ & $ 0.793 \pm 0.03 $ & $ 0.521 \pm 0.06 $\\
    \bottomrule
    \end{tabular}
    }
\end{table*}

% # Average_R = [0.847, 0.823, 0.791, 0.801, 0.757]
% # worst_group_r = [0.489, 0.570, 0.461, 0.610, 0.453]
% \begin{tabular}{l|c|c|c|c|c}
% \hline & Airfoil & NO2 & Exchange-Rate & Electricity & Echo \\
% \hline ERM & $\underline{2.901 \pm 0.067}$ & $0.537 \pm 0.005$ & $0.0236 \pm 0.0031$ & $0.0581 \pm 0.0011$ & $5.402 \pm 0.024$ \\
% mixup & $3.730 \pm 0.190$ & $0.528 \pm 0.005$ & $0.0239 \pm 0.0027$ & $0.0585 \pm 0.0004$ & $5.393 \pm 0.040$ \\
% Mani mixup & $3.063 \pm 0.113$ & $0.522 \pm 0.008$ & $0.0242 \pm 0.0043$ & $0.0583 \pm 0.0004$ & $5.482 \pm 0.066$ \\
% k-Mixup & $2.938 \pm 0.150$ & $0.519 \pm 0.005$ & $0.0236 \pm 0.0029$ & $\underline{0.0575} \pm 0.0002$ & $5.518 \pm 0.034$ \\
% Local Mixup & $3.703 \pm 0.151$ & $\underline{0.517 \pm 0.004}$ & $\underline{0.0236} \pm 0.0024$ & $0.0582 \pm 0.0004$ & $5.652 \pm 0.043$ \\
% MixRL & $3.614 \pm 0.293$ & $0.527 \pm 0.003$ & $0.0238 \pm 0.0037$ & $0.0585 \pm 0.0006$ & $5.618 \pm 0.071$ \\
% \hline C-Mixup (Ours) & $\mathbf{2 . 7 1 7} \pm \mathbf{0 . 0 6 7}$ & $\mathbf{0 . 5 0 9} \pm \mathbf{0 . 0 0 6}$ & $\mathbf{0 . 0 2 0 3} \pm \mathbf{0 . 0 0 1 1}$ & $\mathbf{0 . 0 5 7 0} \pm \mathbf{0 . 0 0 0 6}$ & $\mathbf{5 . 1 7 7} \pm \mathbf{0 . 0 3 6}$
% \end{tabular}

\end{document}